\documentclass{article}

\usepackage{microtype}
\usepackage{graphicx}
\usepackage{subcaption}
\usepackage{booktabs} 

\usepackage{hyperref}

 \usepackage[preprint]{icml2026}

\usepackage{amsmath}
\usepackage{amssymb}
\usepackage{mathtools}
\usepackage{amsthm}

\usepackage{enumitem}
\usepackage{algorithm}
\usepackage[table]{xcolor}
\usepackage{multirow}
\usepackage{makecell}
\usepackage[capitalize,noabbrev]{cleveref}

\theoremstyle{plain}

\theoremstyle{definition}

\theoremstyle{remark}

\usepackage[textsize=tiny]{todonotes}

\icmltitlerunning{When to Lock Attention: Training-Free KV Control in Video Diffusion}

\begin{document}

\twocolumn[
  \icmltitle{When to Lock Attention: Training-Free KV Control in Video Diffusion}



\icmlsetsymbol{equal}{*}
\begin{icmlauthorlist}
\icmlauthor{Tianyi Zeng}{sjtu,equal}
\icmlauthor{Jincheng Gao}{tju,equal}
\icmlauthor{Tianyi Wang}{ut,equal}
\icmlauthor{Zijie Meng}{pku,equal}
\icmlauthor{Miao Zhang}{jmu}
\icmlauthor{Jun Yin}{thu}
\icmlauthor{Haoyuan Sun}{thu}
\icmlauthor{Junfeng Jiao}{ut}
\icmlauthor{Christian Claudel}{ut}
\icmlauthor{Junbo Tan}{thu}
\icmlauthor{Xueqian Wang}{thu}
\end{icmlauthorlist}

\icmlaffiliation{sjtu}{Shanghai Jiao Tong University, Shanghai, China}
\icmlaffiliation{tju}{Tongji University, Shanghai, China}
\icmlaffiliation{ut}{The University of Texas at Austin, Austin, TX, USA}
\icmlaffiliation{pku}{Peking University, Beijing, China}
\icmlaffiliation{jmu}{Jimei University, Xiamen, Fujian, China}
\icmlaffiliation{thu}{Tsinghua University, Beijing, China}

\icmlcorrespondingauthor{Miao Zhang}{zhangmiao@jmu.edu.cn}
  \icmlkeywords{Training-Free Video Editing, Attention Mechanism, Classifier-Free Guidance, Diffusion Hallucination Detection}

  \vskip 0.3in
]



 \printAffiliationsAndNotice{\icmlEqualContribution}

\begin{figure*}[t]
    \centering
    \includegraphics[width=1\linewidth]{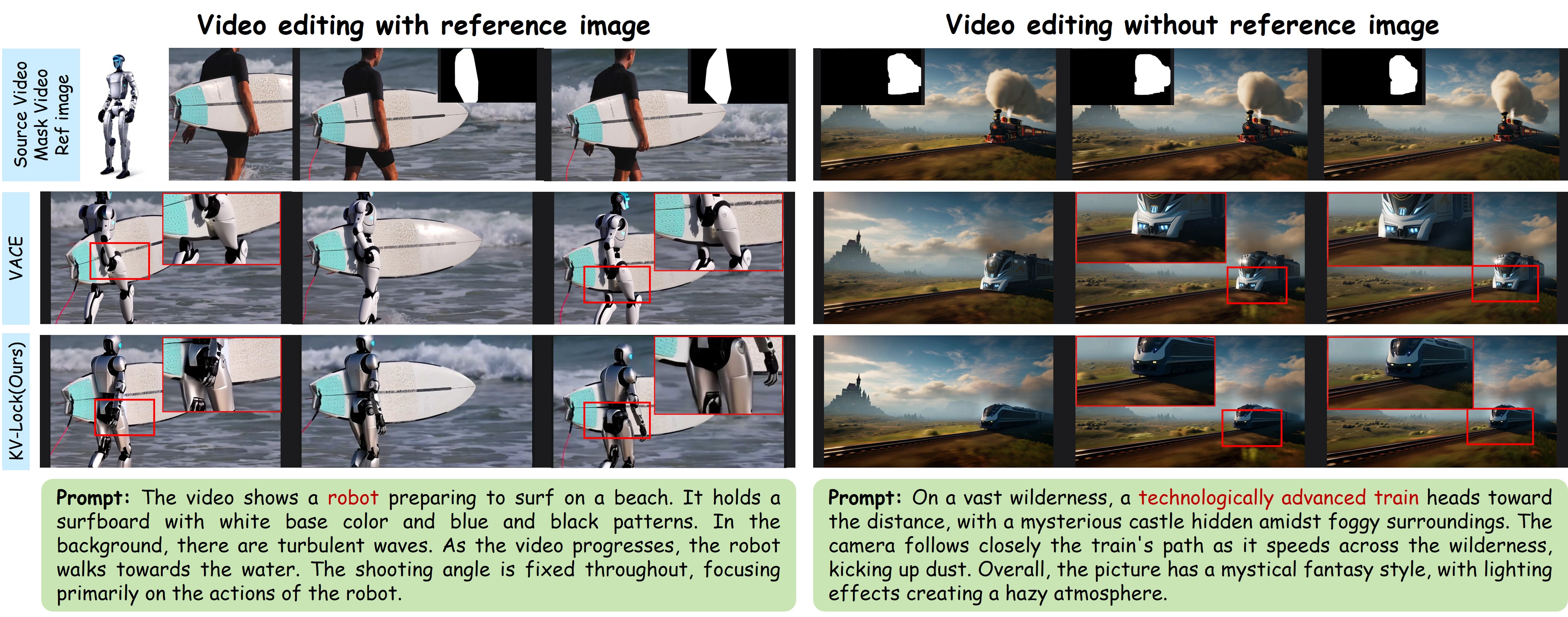}
    \caption{We propose \textbf{KV-Lock}, which dynamically schedules key-value (KV) cache and classifier-free guidance (CFG) based on diffusion model hallucination detection. It enhances foreground quality while ensuring background consistency. Experiments demonstrate that our \textbf{KV-Lock} outperforms VACE \cite{jiang2025vace} in both reference-based and reference-free video editing tasks.}
    \label{fig:main_result}
\end{figure*}

\begin{abstract}
  Maintaining background consistency while enhancing foreground quality remains a core challenge in video editing. 
  Injecting full-image information often leads to background artifacts, whereas rigid background locking severely constrains the model's capacity for foreground generation. 
  To address this issue, we propose KV-Lock, a training-free framework tailored for DiT-based video diffusion models. 
  Our core insight is that the hallucination metric (variance of denoising prediction) directly quantifies generation diversity, which is inherently linked to the classifier-free guidance (CFG) scale.
  Building upon this, KV-Lock leverages diffusion hallucination detection to dynamically schedule two key components: the fusion ratio between cached background key-values (KVs) and newly generated KVs, and the CFG scale. 
  When hallucination risk is detected, KV-Lock strengthens background KV locking and simultaneously amplifies conditional guidance for foreground generation, thereby mitigating artifacts and improving generation fidelity.  
  As a training-free, plug-and-play module, KV-Lock can be easily integrated into any pre-trained DiT-based models. 
  Extensive experiments validate that our method outperforms existing approaches in improved foreground quality with high background fidelity across various video editing tasks.
\end{abstract}

\section{Introduction}
\label{sec:intro}

The advent of text-to-image diffusion models has revolutionized visual content creation, enabling fine-grained control over image synthesis through natural language \cite{saharia2022photorealistic,zhang2023adding,ruiz2023dreambooth}. 
As these models extend to video generation \cite{chen2023control,wu2023tune,guo2024sparsectrl}, a critical new challenge emerges: how can we edit specific objects or regions in an existing video while maintaining the rest of the scene with high fidelity? 
This challenge is particularly acute in professional video production, where even subtle, unintended changes to background elements can break scene continuity and incur substantial post-production costs \cite{zhao2024motiondirector,yin2025slow}.
Unlike image editing \cite{nichol2022glide,zeng2025tcstnet,yin2025prompt}, video editing must satisfy both spatial precision and temporal consistency across dozens of frames, remaining difficult to achieve in practice \cite{ho2022video}.

Current training-based methods, which optimize the model to shorten the diffusion process or reparameterize attention for faster inference, have achieved relatively promising performance when tackling such tasks \cite{meng2023distillation,yin2024one,gao2024matten,chen2025sana,zeng2026pild,zeng2025physics}. 
However, they require substantial computational resources and time when adapting to new data distributions. 
These drawbacks have motivated increasing interest in training-free approaches that improve runtime by reusing latent features or sparse attention computation \cite{wimbauer2024cache,liu2025timestep,zhang2024sageattention,xi2025sparse}.
Among them, inversion-based methods \cite{wallace2023edict} have become an important research direction for applying diffusion models to image and video editing tasks.
A key technique is that DDIM inversion \cite{song2020denoising} can find a noise trajectory that approximately reconstructs the source video when denoised with the original prompt.
Building upon this idea, most of existing approaches \cite{yang2025sparse,zhang2024sageattention2} rely on cross-attention manipulation or interpolation in latent space.
However, these strategies typically provide only coarse control and frequently leak edits into background regions.
Additionally, they often adopt full-image information injection \cite{deng2024fireflow,wang2024taming}, which causes local hallucinations \cite{aithal2024understanding}, especially for target attributes such as color and pose, further limiting their reliability for precise video editing.

Recent studies \cite{zhu2025kv, ouyang2025proedit} have shown that caching key-value (KV) pairs of designated background regions in the DiT architecture \cite{peebles2023scalable} can significantly improve background preservation.
Since attention mechanisms fundamentally determine what each pixel attends to and thus what it becomes, forcing a pixel to attend to cached KVs in the source video should reproduce the original content exactly \cite{qi2023fatezero,liu2024video}. 
However, existing experiments reveal that either fully locking the KV cache or using fixed fusion weights often degrades the foreground generation quality and restricts the model's expressive capability \cite{cai2025ditctrl}.
This raises a deeper and unexplored question: \textit{When should attention be locked to cached KVs, and when should the model be allowed to recompute attention patterns to enable high-quality video editing?}

\begin{figure*}[ht!]
    \centering
    \includegraphics[width=0.9\linewidth]{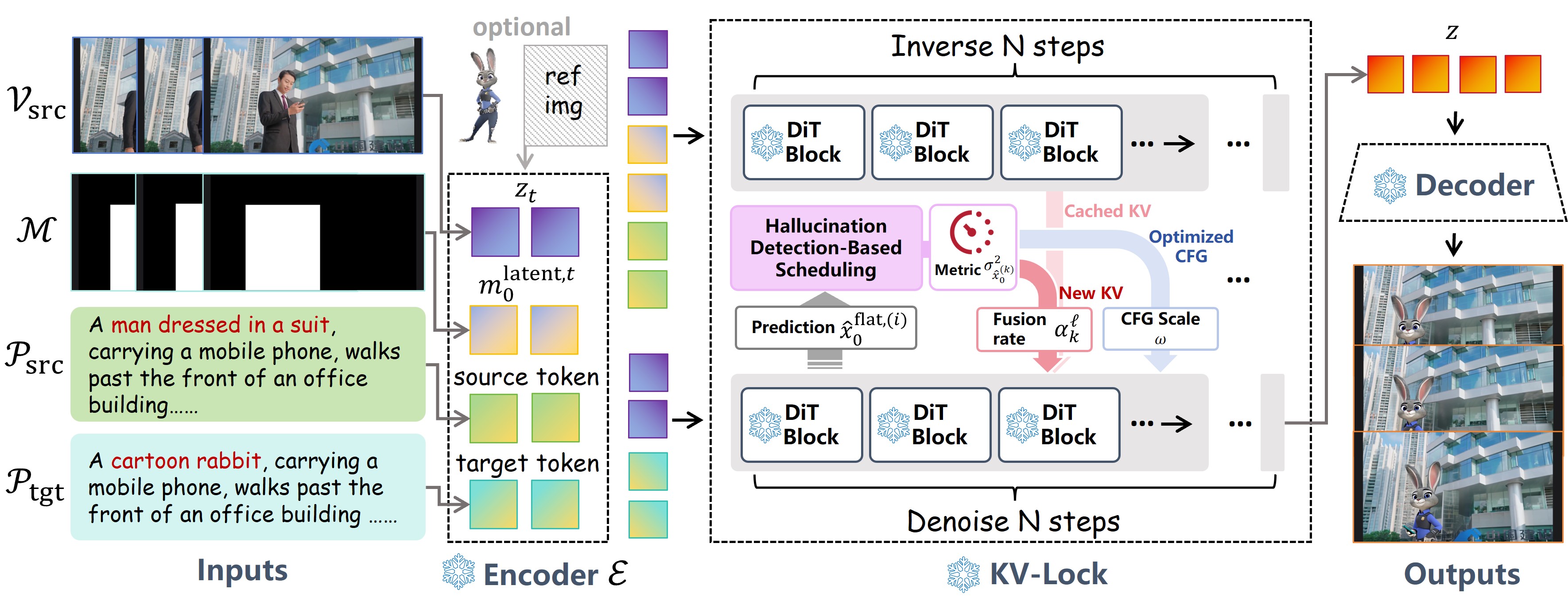}
    \caption{
    Overview of the \textbf{KV-Lock} framework. The encoder first encodes the inputs. Subsequently, during the inversion process, the KV pairs of source tokens are cached. Then, in the denoising process, a hallucination-detection-based scheduler enables the dynamic fusion of newly generated KV and cached KV to ensure background consistency, while dynamic scheduling of CFG enhances foreground quality. Finally, decoding is performed by the decoder.}
    \label{fig:frame}
\end{figure*}

To address this, we propose \textbf{KV-Lock} (Figure \ref{fig:frame}), a hallucination-aware diffusion-based scheduling strategy that dynamically balances foreground quality and background consistency. 
Specifically, KV locking is employed to stabilize background quality, and the injection of background KV values for videos provides a temporal reference for subsequent generation, incidentally addressing the issue of temporal consistency. 
Moreover, the classifier-free guidance (CFG) serves as the foreground optimizer, which is driven by a key observation: the guidance scale in CFG directly regulates the diversity of generated samples, a property inherently linked to the variance used for hallucination detection in diffusion models. 
This natural connection between variance, sample diversity, and CFG scale inspires us to leverage the variance-based hallucination metric to dynamically modulate CFG. 
Collectively, when hallucinations are detected, we increase the weight of cached KV injection to firmly lock the background and ensure its stability, while amplifying the CFG guidance scale to strengthen conditional alignment and thereby mitigate hallucinations. 
This theoretically grounded design realizes dynamic scheduling and elegantly resolves the ``\textit{when to lock attention}'' problem.

The main contributions of this paper are summarized as follows:
\begin{itemize}
    \item We propose a diffusion hallucination detection-based \textbf{KV-Lock} framework, which unifies background control and foreground quality enhancement for video editing by introducing a background KV-locking mechanism together with a foreground-optimized CFG strategy.
    
    \item Through real-time variance-based hallucination detecting, we develop a dynamic scheduling mechanism for \textbf{KV-Lock}, which transforms the question of ``\textit{when to lock attention}'' from heuristic tuning into a principled, variance-driven decision process. This context-aware modulation achieves generalization across diverse video editing scenarios.

    \item Extensive experiments validate the effectiveness of the proposed \textbf{KV-Lock} in achieving high background fidelity and improved foreground quality. As a training-free, plug-and-play module, the proposed method can be seamlessly applied to any other pretrained DiT models.
\end{itemize}

\section{Related Work}

\subsection{Training-Free Text-to-Video Editing via Diffusion Model}

Following the release of Sora \cite{liu2024sora}, which showed the scalability of the DiT architecture for text-to-video generation, several state-of-the-art (SOTA) models, such as CogVideoX \cite{yang2024cogvideox}, Wan \cite{wan2025wan}, and HunyuanVideo \cite{kong2024hunyuanvideo}, have adopted DiT and achieved remarkable performance.
However, these methods typically require extensive training data and computational resources, or rely on reference videos for fine-tuning.
This has motivated growing interest in training-free text-to-video diffusion methods, which have been explored for trajectory control \cite{qiu2024freetraj,chen2025motion,lei2025ditraj}, demographic debiasing \cite{nadeem2025gender,zhong2026fairt2v}, identity preserving \cite{polyak2024movie, he2024id, yuan2025identity}, etc.
In adapting text-to-image models to video editing, FateZero \cite{qi2023fatezero} fused attention maps during inversion and generation process to maintain consistency, while TokenFlow \cite{geyer2023tokenflow} enforced linear combinations of diffusion features based on source correspondences to further improve coherence. 
Slicedit \cite{cohen2024slicedit} enforced temporal consistency by processing spatiotemporal slices of the video volume.
Other studies \cite{gu2024videoswap,wang2025videodirector,li2025flowdirector} leveraged native text-to-video models, and learned spatiotemporal priors for improved temporal consistency.
Despite these advances, existing methods often struggle with substantial visual transformations, resulting in insufficient editing precision and residual background leakage.

\subsection{Key-Value Cache in Video Diffusion Model}

Video diffusion models commonly employ bi-directional attention mechanisms to jointly denoise all video frames \cite{blattmann2023stable}, while this prevents KV cache technique, leading to redundant computation and prohibitive latency for long video sequences.
Although KV caching can reduce KV computation for attention, the attention operation itself still scales quadratically with the KV size, requiring cache eviction strategies to support longer generation \cite{wang2020linformer}. 
To reduce attention complexity without drifting, LongLive \cite{yang2025longlive} introduced a KV sink mechanism that retained key features from earlier frames. 
Block Cascading \cite{bandyopadhyay2025block} shared KV features across parallel blocks to compute self-attention, reducing the reliance external KV caches during inference.
Beyond efficiency, ProEdit \cite{ouyang2025proedit} first employed a mask to decouple attention between background and foreground regions and then inverted the image into noise space while caching KV values of background tokens at each timestep and attention layer.
In contrast to prior methods that rely on simple heuristics, Follow-Your-Shape \cite{long2025follow} presented a trajectory-guided scheduled KV injection scheme to achieve more precise, content-aware control.
These methods highlight the potential of KV-based mechanisms for controllable video editing, but they leave open the question of how to schedule KV usage in a principled, hallucination-aware manner.

\section{Preliminary}

\subsection{Diffusion Hallucination Detection} 

Hallucination in diffusion models is defined as a generated sample that lies entirely outside the support of the true data distribution $q(x)$ \cite{aithal2024understanding}. For a vanishingly small $\epsilon \geq 0$, the hallucination set is 
$H_\epsilon(q) = \{x : q(x) \leq \epsilon\},
$
with the complement $S_\epsilon(q) = \mathbb{X} \setminus H_\epsilon(q)$ being the $\epsilon$-support set of the true distribution. Hallucinations arise from mode interpolation, which means the diffusion model generates samples by smoothly interpolating between nearby valid modes $r,y \in S_\epsilon(q)$ in the representation space, such that the interpolated sample $\theta r + (1-\theta)y \in H_\epsilon(q)$ for $\theta \in [0,1]$.
This interpolation is caused by the neural network’s inability to learn the discontinuous score function of the true data distribution. Instead, the model learns a smoothed approximation of the score function, leading to spurious samples in the low-probability regions between modes.

A critical property of hallucinated samples is the high variance in the trajectory of $\hat{x}_0$ during the reverse process \cite{aithal2024understanding}. This property is quantified by a hallucination detection metric that computes the variance of $\hat{x}_0$ over a range of timesteps $[T_1, T_2]$:
\begin{equation}
\text{Hal}(x) = \frac{1}{|T_2 - T_1|} \sum_{i=T_1}^{T_2} \left(\hat{x}_0^{(i)} - \overline{x_0^{(t)}}\right)^2,
\end{equation}
where $\hat{x}_0^{(i)}$ is the predicted clean sample at timestep $i$ and $\overline{x_0^{(t)}}$ is the mean of $\hat{x}_0^{(i)}$ over $[T_1, T_2]$. A high value of $\text{Hal}(x)$ indicates a hallucinated sample, while a low value indicates an in-support sample.

\subsection{Attention Mechanism as Content Memory}
Video diffusion models based on transformer architectures perform iterative denoising through a sequence of self-attention operations \cite{peebles2023scalable}, which has been widely used in image and video generation \cite{tan2025ominicontrol,yao2025reconstruction,li2024hunyuan,zhang2025tora,yang2024cogvideox}. At each layer $\ell$ and timestep $t_k$, self-attention computes:
$    \text{Attn}(Q, K, V) = \text{softmax}\left(\frac{QK^\top}{\sqrt{d}}\right) V,$
where $Q, K, V \in \mathbb{R}^{N \times d}$ are linear projections of input features with $N$ tokens and $d$ dimensions per attention head. The output at position $i$ is:
\begin{equation}
    o_i = \sum_{j=1}^{N} \underbrace{\text{softmax}\left(\frac{q_i \cdot k_j}{\sqrt{d}}\right)}_{\text{attention weights}} v_j.
    \label{eq:attention_output}
\end{equation}
Equation~\eqref{eq:attention_output} can be interpreted as a differentiable memory retrieval mechanism: query $q_i$ computes similarity scores with all keys $\{k_j\}$, then retrieves a weighted combination of values $\{v_j\}$. Crucially, the output at position $i$ depends on which keys are present and what values they store. If we constrain $K$ and $V$ to specific values while allowing $Q$ to vary, the attention output is restricted to combinations of the fixed value set.

Consider a specific position $i$ where we inject cached KV pairs from the source video: $k_j \gets k_j^{\text{src}}$, $v_j \gets v_j^{\text{src}}$ for all $j$. Even if query $q_i$ differs from the source video's forward pass due to a different denoising trajectory, the attention output becomes:
\begin{equation}
    o_i = \sum_{j=1}^{N} \text{softmax}\left(\frac{q_i \cdot k_j^{\text{src}}}{\sqrt{d}}\right) v_j^{\text{src}},
    \label{eq:locked_attention}
\end{equation}
which is a weighted combination of source-derived values $\{v_j^{\text{src}}\}$. Since these values encode the source video's feature representations at the corresponding noise level and layer depth, the output $o_i$ is constrained to the manifold of source content, providing a deterministic reconstruction mechanism.

\section{Methodology}
\label{sec:methodology}

\subsection{Token-Level KV Cache Locking}
\label{sec:mask_encoding}

\subsubsection{Latent Space Mask Encoding} 

The input video $\mathcal{V}_{\text{src}} \in \mathbb{R}^{3 \times F \times H \times W}$ is encoded by a 3D VAE $\mathcal{E}$ with compression ratio $s = (s_t, s_h, s_w) = (4, 8, 8)$ into latent $z \in \mathbb{R}^{C \times T \times h \times w}$ where $T = \lceil F/s_t \rceil$, $h = H/s_h$, $w = W/s_w$, and $C=16$.
The VAE employs a non-uniform temporal encoding scheme, where the temporal compression operator $\mathcal{E}_t$ acts on disjoint temporal windows of the input video:
\begin{equation}
    z_t = 
    \begin{cases}
        \mathcal{E}_t(\mathcal{V}_0), & t = 0, \\
        \mathcal{E}_t\left(\mathcal{V}_{\left[1 + (t-1)s_t : 1 + ts_t\right]}\right), & t \geq 1,
    \end{cases}
\end{equation}
where $z_t$ denotes the $t$-th temporal slice of the latent tensor $z$, $\mathcal{V}_0$ is the first frame of the input video, and $\mathcal{V}_{\left[1 + (t-1)s_t : 1 + ts_t\right]}$ represents the contiguous subsequence of frames from index $1 + (t-1)s_t$ to $1 + ts_t$ (grouped into windows of size $s_t$) for $t \geq 1$. 
Then, we apply the same temporal aggregation strategy to the mask:
\begin{equation}
    m_0^{\text{latent}, t} = 
    \begin{cases}
        \max\left(\mathcal{M}_0\right), & t = 0, \\
        \max\left(\mathcal{M}_{\left[1 + (t-1)s_t : 1 + ts_t\right]}\right), & t \geq 1,
    \end{cases}
\label{eq:latent_mask_temporal}
\end{equation}
where $m_0^{\text{latent}, t}$ denotes the $t$-th temporal slice of the initial latent mask tensor $m_0^{\text{latent}} \in \{0,1\}^{1 \times T \times h \times w}$, $\mathcal{M}$ is the binary mask. 
The max-pooling operation ensures that if any frame within a temporal window requires editing, the corresponding latent mask slice is marked as $1$.

\subsubsection{Token Space Projection} 

The diffusion transformer processes latents by patchifying spatial dimensions with patch size $p = (p_t, p_h, p_w) = (1, 2, 2)$, yielding $N = T \cdot (h/p_h) \cdot (w/p_w)$ tokens. To project $m_0^{\text{latent}}$ to this token space, we apply 3D max-pooling aligned with the patching operation:
\begin{equation}
\begin{split}
    m_{\text{token}} &= \text{Flatten}\Big(\text{MaxPool3D}(m_0^{\text{latent}}, \\
    &\qquad \text{kernel}=p, \text{stride}=p)\Big) \\
    &\in \{0, 1\}^{N}.
\end{split}
\label{eq:token_mask}
\end{equation}
The resulting binary vector satisfies: $m_{\text{token}}[i] = 1$ if and only if token $i$'s receptive field overlaps at least one masked pixel in the original mask $\mathcal{M}$, ensuring no source content inadvertently enters editing regions due to quantization.

\subsubsection{KV Cache Extraction}

With token-level masks prepared, we extract attention states from the source video that will serve as content anchors during editing.
For each timestep $t_k$ in the denoising schedule $\{t_1, \ldots, t_K\}$, we construct the noisy source latent through forward process:
\begin{equation}
    z_{t_k}^{\text{src}} = \sqrt{\bar{\alpha}_{t_k}}\,\mathcal{E}(\mathcal{V}_{\text{src}}) + \sqrt{1-\bar{\alpha}_{t_k}}\,\epsilon, \quad \epsilon \sim \mathcal{N}(0, I),
    \label{eq:noisy_src}
\end{equation}
where $\bar{\alpha}_{t_k}$ is the cumulative noise schedule coefficient at timestep $t_k$, and $\epsilon$ is sampled once and shared across all timesteps to maintain trajectory consistency. Crucially, we use the \emph{same} $\bar{\alpha}_{t_k}$ schedule as the pretrained model's sampler, ensuring that $z_{t_k}^{\text{src}}$ lies on the model's expected signal-to-noise manifold at every step.
We perform a forward pass
$\hat{\epsilon}_k^{\text{src}} = \epsilon_\theta(z_{t_k}^{\text{src}}, t_k, c_{\text{src}})$
with $c_{\text{src}}$ being the embedded condition from source prompt $\mathcal{P}_{\text{src}}$, and extract KV pairs from all $L{=}24$ transformer blocks:
\begin{equation}
    \mathcal{K}_k^\ell = W_K^{(\ell)} h_{t_k}^{(\ell)}, \quad
    \mathcal{V}_k^\ell = W_V^{(\ell)} h_{t_k}^{(\ell)}, \quad \forall \ell \in \{1, \ldots, L\}.
    \label{eq:kv_cache}
\end{equation}

These cached pairs form a comprehensive memory bank encoding the source video's feature representations at each noise level in each layer's latent space.

The extracted KVs serve as anchors because during editing, the denoising process follows a similar trajectory (same scheduler, same timesteps) but conditioned on target prompt $\mathcal{P}_{\text{tgt}}$. By injecting cached KVs in background regions, we constrain those regions to retrieve features computed from $\mathcal{V}_{\text{src}}$'s forward diffusion, regardless of how foreground regions are changing. 

\subsubsection{Background KV Injection with Hallucination-Based Scheduling}

Since video generation involves denoising the entire frame simultaneously, locking the background constrains the model's ability to generate foreground content, leading to hallucinations in some cases. To address this issue, we introduce a hallucination-aware dynamic fusion rate $\alpha_k \in [0,1]$ for each denoising timestep $t_k$, modulating the KV locking strength according to denoising variance.

The variance $\sigma_{\hat{x}_0^{(k)}}^2$ of $\hat{x}_0$ along the denoising trajectory quantifies hallucination risk \cite{aithal2024understanding}, which is introduced in Section \ref{subsec:hallu}. The fusion rate is given as:
$    \alpha_k = \text{clamp}\left( \frac{\sigma_{\hat{x}_0^{(k)}}^2}{\tau}, 0, 1 \right),$
where $\tau=0.01$ is a hallucination threshold;
$\text{clamp}(\cdot)$ ensures the normalized variance stays within $[0,1]$.
We use the same $\alpha_k$ across all layers, as the variance-based gain already captures the hierarchical hallucination risk in the predicted latent $\hat{x}_0$.

During editing, at each layer $\ell$ and timestep $k$ falling within the final $\kappa=20$ \cite{aithal2024understanding}sampling steps, we intercept the attention computation in the self-attention module and perform token-wise KV interpolation with dynamic scheduling:
\begin{equation}
\begin{aligned}
    Q_k &, K_k^{\text{new}}, V_k^{\text{new}}= W_Q^{(\ell)} h_{t_k}, W_K^{(\ell)} h_{t_k}, W_V^{(\ell)} h_{t_k},\quad \\[4pt]
    K_k^{\text{mix}} &= m_{\text{token}} \odot K_k^{\text{new}} \\
    &\quad + (1 - m_{\text{token}}) \odot 
    \Big( \alpha_k^\ell \cdot \tilde{\mathcal{K}}_k^\ell 
    + (1-\alpha_k^\ell) \cdot K_k^{\text{new}} \Big), \\[4pt]
    V_k^{\text{mix}} &= m_{\text{token}} \odot V_k^{\text{new}} \\
    &\quad + (1 - m_{\text{token}}) \odot 
    \Big( \alpha_k^\ell \cdot \tilde{\mathcal{V}}_k^\ell 
    + (1-\alpha_k^\ell) \cdot V_k^{\text{new}} \Big), \\[4pt]
    \text{Output} &= \text{softmax}\!\left(
    \frac{Q_k (K_k^{\text{mix}})^\top}{\sqrt{d}}\right) 
    V_k^{\text{mix}},
\end{aligned}
\label{eq:kv_injection_dynamic}
\end{equation}
where $m_{\text{token}} \in \{0,1\}^N$ is the token-level mask;
$\tilde{\mathcal{K}}_k^\ell, \tilde{\mathcal{V}}_k^\ell$ are the cached source KV pairs;
$d$ is the dimension of the query/key vectors;
$N$ is the number of spatial tokens in the latent feature map.

\subsection{Foreground Generation Guidance}
\label{sec:pipeline}

\subsubsection{Optimized Scaling Factor for Noise Prediction Correction}
\label{sec:optimized_scale}

Vanilla CFG for diffusion models adopts a fixed guidance scale $\omega$ to linearly interpolate conditional and unconditional noise predictions. 
The preliminary of CFG is provided in Supplementary material Section 1.2.
However, this fixed interpolation fails to compensate for the noise estimation inaccuracies caused by model underfitting—an issue that is particularly prominent in the early denoising stages where the gap between the predicted and true noise is significant. To address this limitation, we introduce an optimizable scalar scaling factor $s \in \mathbb{R}_{>0}$ into the CFG framework, which adaptively corrects the unconditional noise prediction component to reduce the discrepancy between the CFG-guided noise estimate and the true noise of the diffusion process, thus refining the denoising trajectory.

For a given conditioning signal $y$ and its null counterpart $\emptyset$, the model outputs a conditional prediction $\epsilon_\theta(x_t, t|y)$ and an unconditional prediction $\epsilon_\theta(x_t, t|\emptyset)$ at timestep $t$.
We reparameterize the standard CFG-guided noise prediction by integrating the scaling factor $s$ into the unconditional noise term:
\begin{equation}
    \tilde{\epsilon}_{\theta}(x_t, t|y) \triangleq (1-\omega) \cdot s \cdot \epsilon_{\theta}(x_t, t|\emptyset) + \omega \cdot \epsilon_{\theta}(x_t, t|y),
    \label{eq:scaled_cfg_noise}
\end{equation}
where $\omega$ is the original non-negative guidance scale that controls the strength of conditional alignment. 

\subsubsection{Closed-Form Optimization of the Scaling Factor}

Our core objective is to make the scaled CFG-guided noise prediction $\tilde{\epsilon}_{\theta}(x_t, t|y)$ approximate the true noise $\epsilon_t$ of the diffusion process at timestep $t$ :
\begin{equation}
    \mathcal{L}(s) \triangleq \left\| \tilde{\epsilon}_{\theta}(x_t, t|y) - \epsilon_t \right\|_2^2.
    \label{eq:ls_loss_noise}
\end{equation}
A key practical challenge is that the true noise $\epsilon_t$ is not explicitly available for real-world generative tasks. To circumvent this issue, we derive an upper bound of the loss $\mathcal{L}(s)$ using the triangle inequality for vector norms, which eliminates the unobservable true noise $\epsilon_t$ from the optimization objective for $s$:
\begin{equation}
\begin{split}
    \mathcal{L}(s) \leq{}& \left\| \epsilon_{\theta}(x_t, t|y) \right\|_2^2 
    + \left\| \epsilon_t \right\|_2^2 \\
    &+ (1+\omega) \cdot \left\| \epsilon_{\theta}(x_t, t|y) 
    - s \cdot \epsilon_{\theta}(x_t, t|\emptyset) \right\|_2^2.
\end{split}
\label{eq:loss_upper_bound_noise}
\end{equation}
Notably, the first two terms on the right-hand side of Eq.~\ref{eq:loss_upper_bound_noise} are independent of $s$, as they are constant for a given timestep $t$ and model prediction, and thus do not affect the optimization of the scaling factor. Minimizing the upper bound of $\mathcal{L}(s)$ is therefore equivalent to minimizing only the $s$-dependent third term, which simplifies the optimization problem to a classic least-squares fit:
    $\underset{s}{\text{min}}  \left\| \epsilon_{\theta}(x_t, t|y) - s \cdot \epsilon_{\theta}(x_t, t|\emptyset) \right\|_2^2.$
This problem has a closed-form analytical solution derived by taking the derivative of the objective with respect to $s$, setting the derivative to zero, and solving for $s$ \cite{fan2025cfg}. The resulting optimal scaling factor $s^*$ is given by:
\begin{equation}
    s^* = \frac{\left\langle \epsilon_{\theta}(x_t, t|y), \epsilon_{\theta}(x_t, t|\emptyset) \right\rangle}{\left\| \epsilon_{\theta}(x_t, t|\emptyset) \right\|_2^2+\varepsilon},
    \label{eq:opt_scaling_factor_noise}
\end{equation}
where $\langle \cdot, \cdot \rangle$ denotes the vector inner product, $\varepsilon$ is introduced to avoid division by zero.
Geometrically, the optimal scaling factor $s^*$ corresponds to the orthogonal projection of the conditional noise prediction vector $\epsilon_{\theta}(x_t, t|y)$ onto the unconditional noise prediction vector $\epsilon_{\theta}(x_t, t|\emptyset)$. This projection ensures that the scaled unconditional noise prediction $s^* \cdot \epsilon_{\theta}(x_t, t|\emptyset)$ is the closest vector to the conditional noise prediction along the direction of the unconditional output, effectively aligning the two noise estimates in the latent space and reducing the bias introduced by model underfitting.

\subsubsection{Hallucination-Aware Dynamic Guidance Scale}
\label{sec:dynamic_cfg_hallucination}

To mitigate hallucinations in foreground generation  at final $\kappa$ timesteps (further details on hallucination detection are provided in Subsection~\ref{subsec:hallu}), we introduce a dynamic adjustment to the CFG guidance scale $\omega$ that amplifies conditional alignment strength when hallucination risk is detected.

For each denoising step $k$ within the window $W=10$, whose rationality has been validated through extensive experiments, if hallucination risk is detected, we adjust:
$    \omega =  \omega_0 \cdot \text{clamp}\left(\frac{\sigma_{\hat{x}_0^{(k)}}^2}{\tau}, 0, b\right),$
where $\sigma_{\hat{x}_0^{(k)}}^2$ is the variance of predicted $x_0$ at timestep $k$ along the denoising trajectory;
$\tau=0.01$ is the hallucination threshold;
$b=2$ is the clamp boundary, both values validated by extensive experiments and with reference to \cite{aithal2024understanding};
If no hallucination risk is detected ($\sigma_{\hat{x}_0^{(k)}}^2 < \tau$), or if the step falls without the final $\kappa=20$ \cite{aithal2024understanding}sampling steps, the dynamic guidance scale remains unchanged at the base value $\omega_0$. 

The guidance scale $\omega$ of CFG regulates the diversity of generated samples. During the late denoising steps of hallucination-prone samples, the diffusion trajectory tends to exhibit significant fluctuations; dynamically increasing $\omega$ thus helps constrain sample diversity and ensure the stability of the diffusion process. 
In contrast, during the early stages of diffusion, all samples exhibit high variance, making such adaptive scheduling unnecessary. 
Our core insight lies in the fact that this hallucination-aware scheduler for diffusion models is perfectly aligned with the inherent characteristics of CFG, leveraging the scale’s diversity-regulation property to mitigate trajectory instability only when needed.

\begin{table*}[ht]
\begin{center}

\setlength{\tabcolsep}{2.1pt} 

\caption{Quantitative results.}
\resizebox{\linewidth}{!}{

\begin{tabular}{l|cccccc|cc|cccc|c}
\toprule
\multirow{2}*{\textbf{Method}} & \multicolumn{6}{c|}{\textbf{VBench Metrics}} & \multicolumn{2}{c|}{\textbf{Background}} & \multicolumn{4}{c|}{\textbf{User Study}} & \multirow{2}*{\makecell{\textbf{Inf.} \\ \textbf{Time}}$\downarrow$}\\  
~ & SC$\uparrow$ & BC$\uparrow$ & MS$\uparrow$ & AQ$\uparrow$ & IQ$\uparrow$ & Ave.$\uparrow$ & {SSIM$\uparrow$} & {PSNR$\uparrow$} & PF$\uparrow$ & FC$\uparrow$ & VQ$\uparrow$ & \makecell{Ave.$\uparrow$}&~ \\  

\midrule

FateZero\cite{qi2023fatezero} &  $87.17\%$&  $92.89\%$ & $94.72\%$&  $53.84\%$ & $57.53\%$ & $77.23\%$ &$0.7151$&$17.57$ &$2.04$ &$1.78$ & $1.41$ & $1.74$ & \underline{$3.98$}\\

FLATTEN\cite{cong2023flatten} &$92.90\%$ & $95.54\%$&$97.48\%$ & $53.24\%$  & $59.41\%$ & $79.71\%$ & $0.7716$ & $19.30$& {$2.41$} &$3.15$ & $2.26$ & $2.60$ & $\boldsymbol{1.14}$ \\

TokenFlow\cite{geyer2023tokenflow} & ${93.64\%}$  & $96.17\%$ & $98.46\%$ & $57.22\%$  & $69.67\%$ & $83.03\%$ & $0.8050$ &$20.07$ & $2.00$ & $2.78$ & $2.74$ & $2.51$ & $11.92$\\

CFG-Zero*\cite{fan2025cfg}& $93.80\%$ & $95.99\%$ & \underline{${98.75\%}$} & $61.22\%$  &$71.04\%$ & $84.16\%$ &$0.9107$ &$26.65$ & $4.30$ & $3.93$ & $3.81$ & $4.01$ & $5.58$\\

APG \cite{sadat2024eliminating} & $93.39\%$ & \underline{$96.25\%$} & $\boldsymbol{98.83\%}$ & $60.09\%$  &$71.53\%$ & $84.02\%$ &$0.9211$ &$26.04$ & $4.22$ & $3.85$ & $3.78$ & $3.95$ & $5.80$\\

ProEdit\cite{ouyang2025proedit}& \underline{$93.96\%$} & {$96.23\%$} & $98.56\%$ & \underline{$61.62\%$}  &$\boldsymbol{72.23\%}$ & \underline{$84.52\%$} &$0.9116$ &$27.57$ & $4.37$ & $3.78$ & $\boldsymbol{4.04}$ & $4.06$ & $7.20$\\

VACE \cite{jiang2025vace} & $93.82\%$ & $95.85\%$ & ${98.74\%}$ & $61.25\%$  & $71.01\%$ & $84.13\%$ & \underline{$0.9218$} & $\boldsymbol{31.20}$& \underline{${4.41}$} & \underline{${4.11}$} & $3.78$& \underline{$4.10$} & $5.25$\\

\rowcolor{gray!20}\textbf{KV-Lock (Ours)}  &  $\boldsymbol{94.56\%}$ & $\boldsymbol{96.92\%}$ &  {$98.57\%$} & $\boldsymbol{62.15\%}$  & \underline{$72.18\%$} & $\boldsymbol{84.87\%}$  & $\boldsymbol{0.9309}$ & \underline{$31.04$} & $\boldsymbol{4.48}$ & $\boldsymbol{4.30}$ & \underline{$3.85$} & $\boldsymbol{4.21}$& $7.39$ \\ \bottomrule

\end{tabular}
}
\label{table:result}
\end{center}
\end{table*}

\begin{table*}[ht]
\begin{center}

\setlength{\tabcolsep}{3pt} 

\caption{Results of ablation study.}
\resizebox{0.8\linewidth}{!}{

\begin{tabular}{l|cccccc|cc}
\toprule
\multirow{2}*{\textbf{Method}} & \multicolumn{6}{c|}{\textbf{VBench Metrics}} & \multicolumn{2}{c}{\textbf{Background}}\\  
~ & SC$\uparrow$ & BC$\uparrow$ & MS$\uparrow$ & AQ$\uparrow$ & IQ$\uparrow$ & Ave.$\uparrow$ & {SSIM$\uparrow$} & {PSNR$\uparrow$} \\  

\midrule

Variance KV schedule only &  $93.01\%$&  \underline{$95.89\%$} & $98.10\%$&  $60.28\%$ & $71.19\%$ & $83.69\%$ &$0.9129$&\underline{$31.01$}\\

CFG $\omega$ schedule only&  \underline{$93.32\%$}&  $93.89\%$ & $97.72\%$&  $61.84\%$ & $70.53\%$ & $83.46\%$ &$0.9217$&$29.84$\\

CFG $s^*$ schedule only&  $91.76\%$&  $92.18\%$ & $96.92\%$&  $60.16\%$ & $70.17\%$ & $82.24\%$ &$0.9141$&$29.59$\\

CFG $s^*$ and $\omega$ schedule &  $93.28\%$&  $95.71\%$ & $\boldsymbol{98.63\%}$&  $61.67\%$ & $70.95\%$ & \underline{$84.05\%$} &$0.9130$&$30.55$\\

Variance KV and CFG $\omega$ schedule &  $93.26\%$&  $94.79\%$ & $97.88\%$&  $61.70\%$ & \underline{$71.25\%$} & $83.78\%$ &\underline{$0.9261$}&$30.53$\\

Variance KV and CFG $s^*$ schedule &  $92.98\%$&  $95.87\%$ & ${98.11\%}$&  \underline{$62.01\%$} & $70.94\%$ & $83.98\%$ &$0.9258$&$30.72$\\

Fixed fusion $\alpha_k=0.5$  &  $90.33\%$&  $93.97\%$ & $97.51\%$&  $60.93\%$ & $70.18\%$ & $82.58\%$ &$0.9175$&$30.90$\\

Global hallucination detection  &  $93.14\%$&  $95.85\%$ & $98.28\%$&  $61.90\%$ & $71.09\%$ & \underline{$84.05\%$} &$0.9254$&$30.96$\\

\rowcolor{gray!20}\textbf{Full Model}  &  $\boldsymbol{94.56\%}$ & $\boldsymbol{96.92\%}$ &  \underline{$98.57\%$} & $\boldsymbol{62.15\%}$  & {$\boldsymbol{72.18\%}$} & $\boldsymbol{84.87\%}$  & $\boldsymbol{0.9309}$ & $\boldsymbol{31.04}$ \\ \bottomrule

\end{tabular}
}
\label{table:ablation}
\end{center}
\end{table*}

\subsection{When to Lock Attention}
\label{subsec:hallu}

A key theoretical insight from diffusion model research justifies hallucination as a scheduling cue: \textit{the trajectory of the predicted clean latent $\hat{x}_0$ directly reflects generation stability} \cite{aithal2024understanding}. For in-support samples, $\hat{x}_0$ converges to a consistent representation in late denoising stages, resulting in low trajectory variance; for hallucinated samples, $\hat{x}_0$ fluctuates due to the model’s uncertainty in mode interpolation regions, leading to high variance. 
However, computing the variance over the entire frame significantly weakens the hallucination signal. We therefore compute the local variance, whose effectiveness has also been verified in ablation studies.
This local variance thus serves as an observable proxy for hallucination risk, enabling training-free, real-time scheduling of KV locking strength.

We compute the hallucination metric by tracking the local variance of the masked $\hat{x}_0$ during reverse diffusion. 
The implementation leverages a sliding window tracker to capture temporal dynamics without excessive memory overhead.

For each denoising timestep $t_k$, we extract the predicted clean latent $\hat{x}_0^{(k)}$ from the scheduler’s output. 
To increase the amplitude of the hallucination signal in the foreground region, we first flatten $\hat{x}_0^{(k)}$ into a 1D vector, then compute the masked batch-wise mean to aggregate statistical characteristics as 
$\hat{x}_0^{\text{masked}, (k)} = \frac{1}{B} \sum_{b=1}^B {\text{Flatten}\left(\hat{x}_0^{(k, b)} \odot m_0^{\text{latent}}\right)}$
where $B$ is the batch size, and $\text{Flatten}(\cdot)$ reshapes the multi-dimensional latent tensor into a 1D vector.

We then maintain a sliding window of the most recent $W$ $\hat{x}_0^{\text{masked}, (k)}$ vectors. The variance of the trajectory within the window is computed as:
   \begin{equation}
       \sigma_{\hat{x}_0^{(k)}}^2 = \frac{1}{W-1} \sum_{i=t_k-W+1}^{t_k} \left( \hat{x}_0^{\text{masked}, (i)} - \bar{\hat{x}}_0^{\text{masked}} \right)^2,
       \label{eq:x0_variance}
   \end{equation}
where $\bar{\hat{x}}_0^{\text{masked}} = \frac{1}{W} \sum_{i=t_k-W+1}^{t_k} \hat{x}_0^{\text{masked}, (i)}$ is the window-wise mean of the masked flattened $\hat{x}_0$ vectors. 
A hallucination risk is flagged if $\sigma_{\hat{x}_0^{(k)}}^2 > \tau$.

\begin{figure*}[ht!]
    \centering
    \includegraphics[width=\linewidth]{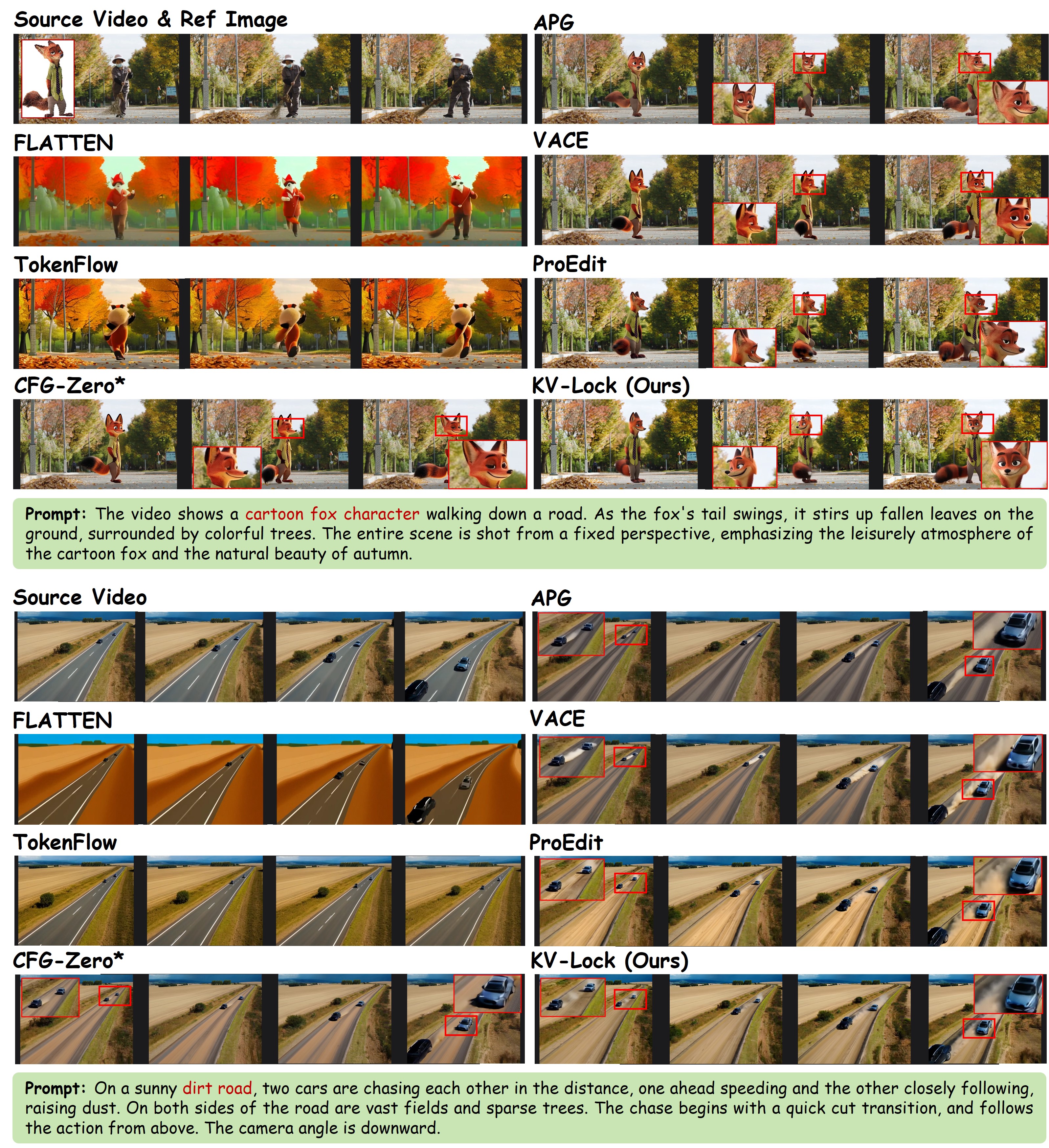}
    \caption{Two comparison experimental samples. It can be observed that FLATTEN \cite{cong2023flatten} and TokenFlow \cite{geyer2023tokenflow} achieve an almost complete failure, with extensive distortions and artifacts. Specifically, in the first sample, the two eyes of the fox generated by VACE \cite{jiang2025vace} and CFG-Zero* \cite{fan2025cfg} are asymmetrical and unnatural, APG \cite{sadat2024eliminating} causes obvious distortion to the fox. while the fur texture rendered by \textbf{KV-Lock} is more refined than that of ProEdit \cite{ouyang2025proedit}. In the second sample, the distant dust generated by VACE \cite{jiang2025vace} is excessively bright and unrealistic, and the road surface is an asphalt road with soil instead of a dirt road, also appearing in CFG-Zero* \cite{fan2025cfg} and APG \cite{sadat2024eliminating}; 
    In CFG-Zero* \cite{fan2025cfg}, the dust raised by the front vehicle exhibits an obvious unnatural boundary. For APG \cite{sadat2024eliminating}, the rear vehicle generates no dust at all, and ProEdit \cite{ouyang2025proedit} produces more prominent distant dust than near dust, which are contrary to common sense.}
    \label{fig:result2}
\end{figure*}

\begin{figure*}[t!]
    \centering
    \includegraphics[width=\linewidth]{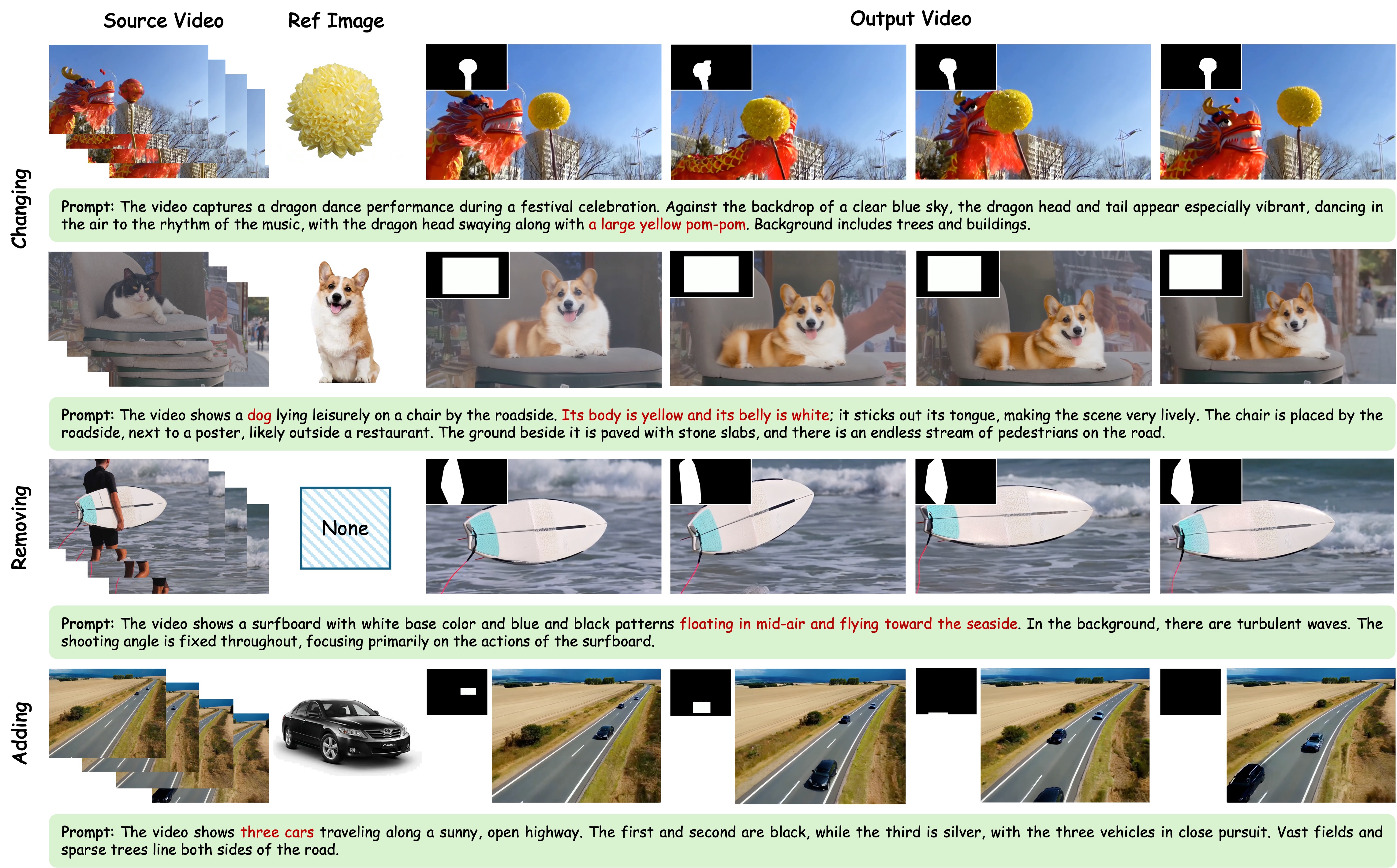}
    \caption{More samples of KV-Lock, including changing, removing and adding tasks.}
    \label{fig:result3}
\end{figure*}

\section{Experiments}
\subsection{Setup}

This work is developed based on the VACE \cite{jiang2025vace}.
The model is deployed on an A100 80GB GPU.
We select training-free video editing methods including FateZero\cite{qi2023fatezero}, FLATTEN\cite{cong2023flatten}, TokenFlow\cite{geyer2023tokenflow}, 
CFG-Zero*\cite{fan2025cfg},
APG\cite{sadat2024eliminating},
ProEdit\cite{ouyang2025proedit}
and training-based VACE\cite{jiang2025vace} for comparison.
Notably, Wan \cite{wan2025wan} is one of the current mainstream SOTA models. Thus, CFG-Zero*, APG, ProEdit, and KV-Lock are all deployed on Wan 2.1. Since the remaining three methods do not have open-source Wan implementations, FateZero, FLATTEN, and Tokenflow use StableDiffusion 2.1 \cite{rombach2022high} as base model.
The hyperparameter settings and baseline details are provided in Supplementary material Section 2.1 and 2.2.

The test dataset we used 22 scenarios of VACE-Benchmark \cite{jiang2025vace} and 30 videos collected from the Internet, 52 samples in total. The number of frames per video ranges from 80 to 210, with resolutions of 480×832.

We selected all metrics for self-built dataset provided by VBench \cite{huang2024vbench} including Subject consistency (SC), Background consistency (BC), Motion smoothness (MS), Aesthetic quality (AQ) and Imaging quality (IQ); and SSIM, PSNR; and user study metrics including Prompt following (PF), Frame consistency (FC), Video quality (VQ) provided by VACE \cite{jiang2025vace}.
It should be noted that the SC and BC metrics in VBench measure consistency with the prompt rather than with the original video. We therefore additionally introduce the SSIM and PSNR on background to quantify the absolute value of background consistency.

\subsection{Results}

Quantitative experimental results are presented in Table \ref{table:result}. 
It can be observed that KV-Lock achieves the best overall performance on VBench benchmarks, achieving promising results on SC, BC, and AQ. The two CFG-based optimization methods, CFG-Zero* and APG, perform favorably on the MS via guidance control. Meanwhile, ProEdit, a KV-editing approach based on fixed background KV fusion, also achieves competitive performance across multiple metrics.
For background quantitative metrics, VACE attains the highest PSNR score due to its strong modeling capability, which results in less noise in the reconstructed background. By contrast, KV-Lock achieves superior performance on SSIM through the dynamic background locking mechanism we proposed.
Two comparison samples are selected for visualization in Figure \ref{fig:result2}, more samples are provided in Figure \ref{fig:result3} and  Supplementary material Section 2.4.

In addition, we report the runtime in Table \ref{table:result}. Since the number of frames varies per sample and the data loading pipelines of different baselines differ, we select 100-frame samples and report the per-iteration diffusion sampling time (s) for a fair comparison. 
The results show that our method is slower than other methods due to KV caching and sliding-window computation, which is a limitation to be addressed in future work. 
KV caching requires approximately 10GB of GPU memory on average, whose detailed memory footprint is provided in Supplementary material Section 2.3.
Nevertheless, the trade-off between running cost and generation quality is acceptable overall.

\begin{figure*}[t!]
    \centering
    \includegraphics[width=\linewidth]{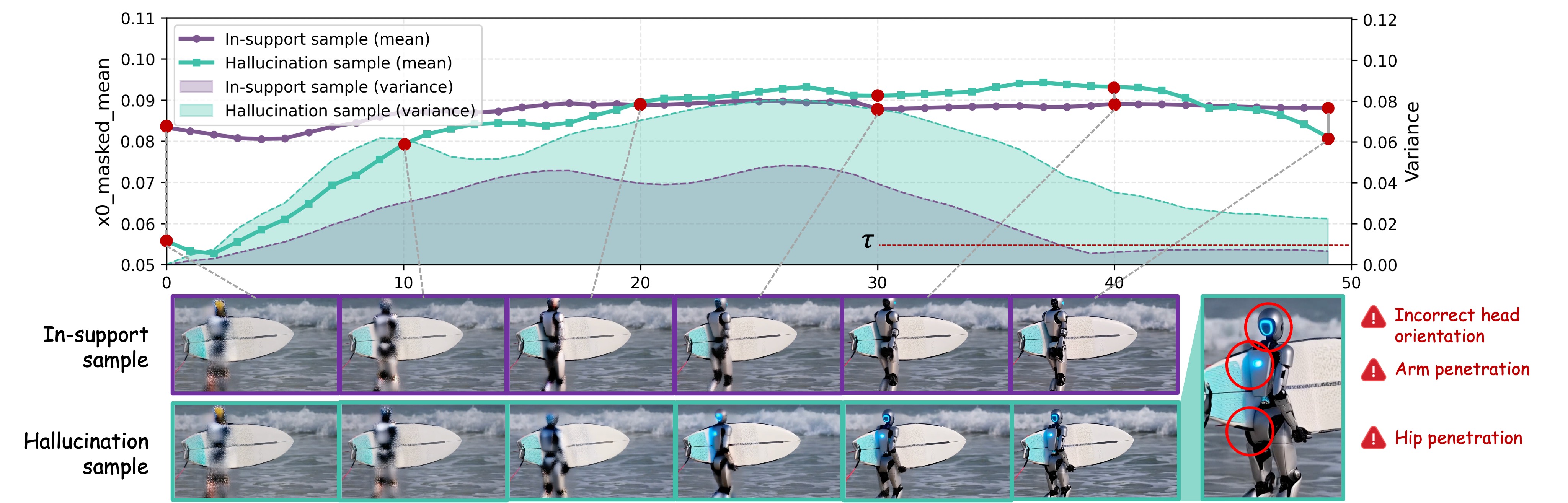}
    \caption{In the early diffusion stage, the variance of in-support and hallucination samples are both high. After hallucination is detected, KV-Lock control the variance under threshold $\tau$ through dynamic scheduling.}
    \label{fig:hallu_ablation}
\end{figure*}

\subsection{Ablation Study}

Ablation study on the core modules and hyperparameters are conducted, with the results presented in Table \ref{table:ablation}.
Experiments validate the effectiveness of hallucination based scheduling and demonstrate the performance gain brought by each individual module.
Furthermore, we test a fixed fusion $\alpha_k$ and find that it cannot adapt to variations in the diffusion path, leading to degraded performance. 
We also evaluate global hallucination detection, which validates our earlier intuition: global detecting weakens the hallucination signal and resulting in missed detections and inferior performance.

Furthermore, we demonstrate the effectiveness of hallucination detection, with the results presented in Figure \ref{fig:hallu_ablation}. 
It can be observed that in the latter stage of the denoising trajectory, paths where the variance decreases and stabilizes ultimately yield high-quality generation results. 
If hallucinations occur, the scheduler intervenes in a timely manner to regulate the quality of the generation.

\subsection{User Study}

We distributed 70 questionnaires and ultimately collected 54 valid responses. The user study was conducted across three dimensions: prompt following, frame consistency, and video quality. Participants were asked to rate on a scale of 1 to 5. The results are also presented in Table \ref{table:result}, which shows that our method has received human visual approval for its comprehensive generation capability.

\section{Conclusions}

This paper proposes KV-Lock, a KV control framework for the DiT architecture based on diffusion model hallucination detection, designed to enhance foreground generation quality while ensuring background consistency. 
Extensive experiments verify the feasibility of our novel idea for video generation quality control based on hallucination detection, and demonstrate promising performance across multiple metrics.
This work also leaves an open question: hallucination in diffusion models lacks a fixed definition, meaning it can be detected not only by variance but also via numerous alternative technical approaches.
Furthermore, our approach currently relies on existing mask inputs to separate foreground and background, and KV caching requires a pre-run pass which leads to relatively more inference time. 
More concise input modalities and optimized caching strategies will be the focus of future work.

\bibliography{example_paper}
\bibliographystyle{icml2026}

\newpage
\appendix
\onecolumn
\section{Preliminary}
\subsection{Diffusion Models}

Let $q(x)$ denote the true data distribution of the training samples. Diffusion models consist of a forward noising process and a reverse denoising process, with discrete timesteps $t = 1,2,\dots,T$. In the forward process, Gaussian noise is incrementally added to the clean sample $x_0 \sim q(x)$, yielding a noisy sample $x_t$ at timestep $t$. The transition kernel is defined as:
\begin{equation}
q(x_t | x_{t-1}) = \mathcal{N}\left(\sqrt{1-\beta_t} x_{t-1}, \beta_t I\right),
\end{equation}
where $\beta_t \in (0,1)$ is the noise schedule at timestep $t$. The marginal distribution of $x_t$ given $x_0$ has a closed-form:
$
q(x_t | x_0) = \mathcal{N}\left(\sqrt{\bar{\alpha}_t} x_0, (1-\bar{\alpha}_t) I\right),
$
with $\alpha_t = 1-\beta_t$ and $\bar{\alpha}_t = \prod_{j=1}^t \alpha_j$ denoting the cumulative product of $\alpha_j$. For sufficiently large $T$, $x_T$ is approximately standard Gaussian noise $\mathcal{N}(0, I)$.

The reverse process learns to recover clean samples from noisy ones by modeling the distribution $p_\theta(x_{t-1} | x_t)$ with a neural network parameterized by $\theta$:
\begin{equation}
p_\theta(x_{t-1} | x_t) = \mathcal{N}\left(\mu_\theta(x_t, t), \Sigma_\theta(x_t, t)\right),
\end{equation}
where the mean $\mu_\theta(x_t, t)$ is derived from the predicted noise $\epsilon_\theta(x_t, t)$:
\begin{equation}
\mu_\theta(x_t, t) = \frac{1}{\sqrt{\alpha_t}}\left(x_t - \frac{1-\alpha_t}{\sqrt{1-\bar{\alpha}_t}} \epsilon_\theta(x_t, t)\right).
\end{equation}
The predicted clean sample $\hat{x}_0$ is computed as:
$
\hat{x}_0 = \frac{1}{\sqrt{\bar{\alpha}_t}}\left(x_t - \sqrt{1-\bar{\alpha}_t} \epsilon_\theta(x_t, t)\right).
$

\subsection{Classifier-Free Guidance}

CFG is a technique for enhancing the fidelity and conditional alignment of flow matching and diffusion models during sampling, which eliminates the need for pre-trained classifiers by jointly learning conditional and unconditional velocity or noise prediction branches within a single model \cite{ho2022classifier, zheng2023guided}. 
For a given conditioning signal $y$ and its null counterpart $\emptyset$, the model outputs a conditional prediction $\epsilon_\theta(x_t, t|y)$ and an unconditional prediction $\epsilon_\theta(x_t, t|\emptyset)$ at timestep $t$.

The guided noise prediction for CFG is formulated as a linear interpolation between the unconditional and conditional outputs, weighted by a non-negative guidance scale $\omega$ that controls the strength of conditioning:
\begin{equation}
\hat{\epsilon}_\theta(x_t, t|y) \triangleq (1-\omega) \cdot \epsilon_\theta(x_t, t|\emptyset) + \omega \cdot \epsilon_\theta(x_t, t|y).
\end{equation}
This guided noise $\hat{\epsilon}_\theta(x_t, t|y)$ then replaces the original conditional noise in the diffusion model's mean update formula for the reverse denoising process, yielding the CFG-guided mean:
\begin{equation}
\hat{\mu}_\theta(x_t, t|y) = \frac{1}{\sqrt{\alpha_t}}\left(x_t - \frac{1-\alpha_t}{\sqrt{1-\bar{\alpha}_t}} \hat{\epsilon}_\theta(x_t, t|y)\right).
\end{equation}
When $\omega=1$, the guidance is deactivated, and the model uses only the conditional prediction $\epsilon_\theta(x_t, t|y)$ for sampling. For $\omega>1$, the model amplifies the conditional signal to enforce stronger alignment with the input $y$, while $\omega<1$ weakens the conditioning by blending in the unconditional distribution.

\section{Experiments}

\subsection{Implementation Details}

Our method is developed based on the VACE architecture, with no modifications to the original VACE model parameters \cite{jiang2025vace}.
The hyperparameter for KV-Lock are presented in Table \ref{table:hyper}.

\begin{table*}[h!]
\begin{center}

\setlength{\tabcolsep}{10pt} 

\caption{Hyperparameter Settings.}
\resizebox{0.75\linewidth}{!}{

\begin{tabular}{lcl}
\toprule

\textbf{Hyperparameter} & \textbf{Value}& \textbf{Note} \\  

\midrule

$\tau$ &  $0.01$ & Hallucination threshold\\

Total diffusion timestep &  $50$ & -\\

$W$ &  $10$ & Hallucination detection window size \\

$\kappa$ & $20$ & Hallucination detection final step \cite{aithal2024understanding}  \\

$\omega_0$ & $5$ & Origin CFG scale \cite{jiang2025vace}  \\

$b$ & $2$ & Clamp boundar  \\

\bottomrule

\end{tabular}
}
\label{table:hyper}
\end{center}
\end{table*}

\begin{figure*}[ht]
    \centering
    \includegraphics[width=0.55\linewidth]{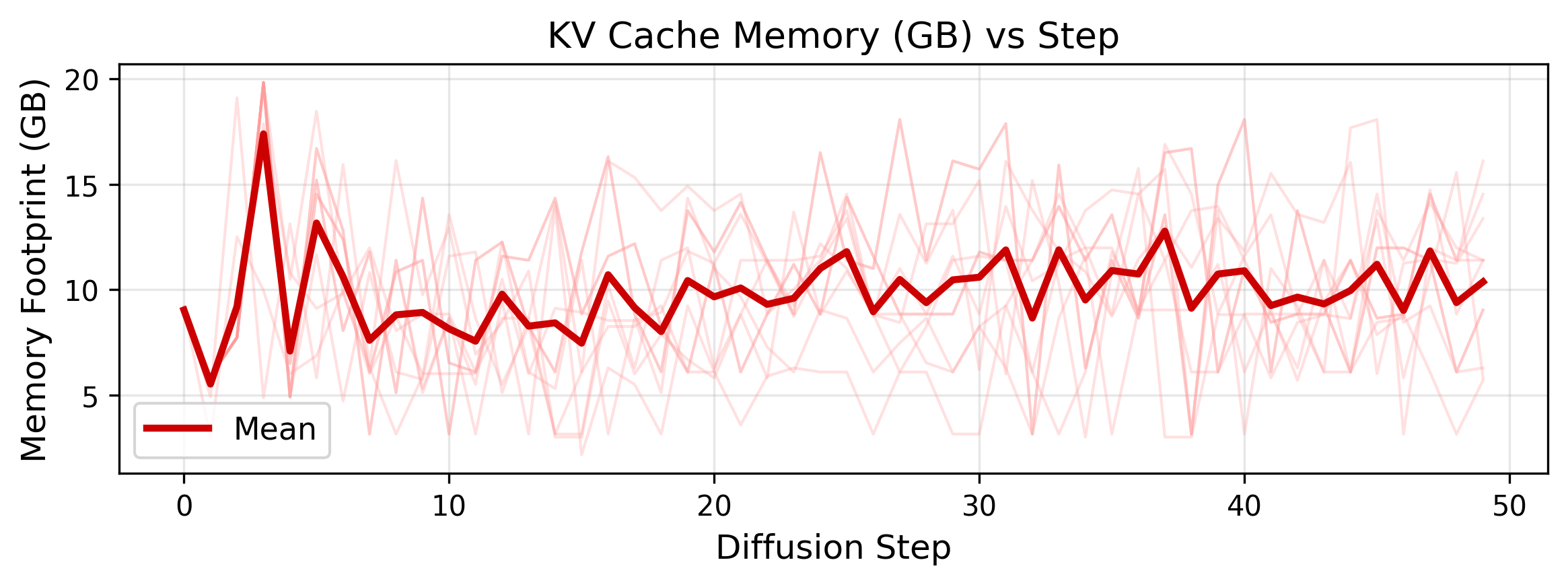}
    \caption{KV-cache memory footprint for each sampling timestep.}
    \label{fig:memory}
\end{figure*}

\subsection{Baseline Details}

We introduce the details of baselines we selected, covering both training-based and training-free mainstream video generation methods.

The following three methods use StableDiffusion as base model.
\textbf{FateZero} \cite{qi2023fatezero} proposes a zero-shot text-based editing method on real-world videos without per-prompt training or use-specific mask. The method is the first one to show the ability of zero-shot text-driven video style and local attribute editing from the trained text-to-image model. 
\textbf{FLATTEN} \cite{cong2023flatten} proposes an optical-flow-based approach that incorporates optical flow into the attention module to address text inconsistency issues.
\textbf{TokenFlow} \cite{geyer2023tokenflow} enforces text-video consistency in the feature space by propagating inter-frame features within the video.
Notably, the above three methods rely on neither mask inputs nor reference images. Consequently, they cannot preserve background consistency during full-image processing, resulting in severe corruption of global content. Thus, they fail to achieve satisfactory performance in most scenarios.

Several other methods use Wan as their base model.
\textbf{VACE} \cite{jiang2025vace} serves as the foundational architecture for subsequent training-free methods. It proposes an all-in-one framework that enables multi-task video editing and achieves SOTA performance across multiple metrics, with its core architecture built upon DiT.
\textbf{APG} \cite{sadat2024eliminating} is an optimization approach for CFG scale, which improves the realism of generated videos by decomposing the scale into parallel and perpendicular components to separately control saturation and details.
\textbf{CFG-Zero*} \cite{fan2025cfg} is an optimized CGF ensemble strategy that achieves better conditional control through decoupling the CFG scale and zero-initialization of the diffusion process.
\textbf{ProEdit} \cite{ouyang2025proedit} improves video generation quality by fusing the background key-value (KV) features with newly generated KV features using a fixed fusion ratio.

All these methods aim to achieve better text-image consistency and generation quality, with focuses ranging from system pipeline design, CFG optimization, to KV optimization. They cover the research areas involved in the method proposed in this paper, ensuring a sufficiently comprehensive comparison.

\subsection{Memory Footprint}

Our method requires an initial forward pass over the raw video for KV caching during sampling, which introduces extra GPU memory consumption. Thus, we report the additional memory overhead caused by KV caching here.
Since the memory usage varies significantly across samples along the sampling path, the average GPU memory consumption is plotted with the dark red line in Figure \ref{fig:memory}.
The average GPU memory consumption is approximately 10GB, with a maximum of around 20GB and a minimum of about 3GB.

\subsection{More Results}

\begin{figure*}[htpb!]
    \centering
    \includegraphics[width=\linewidth]{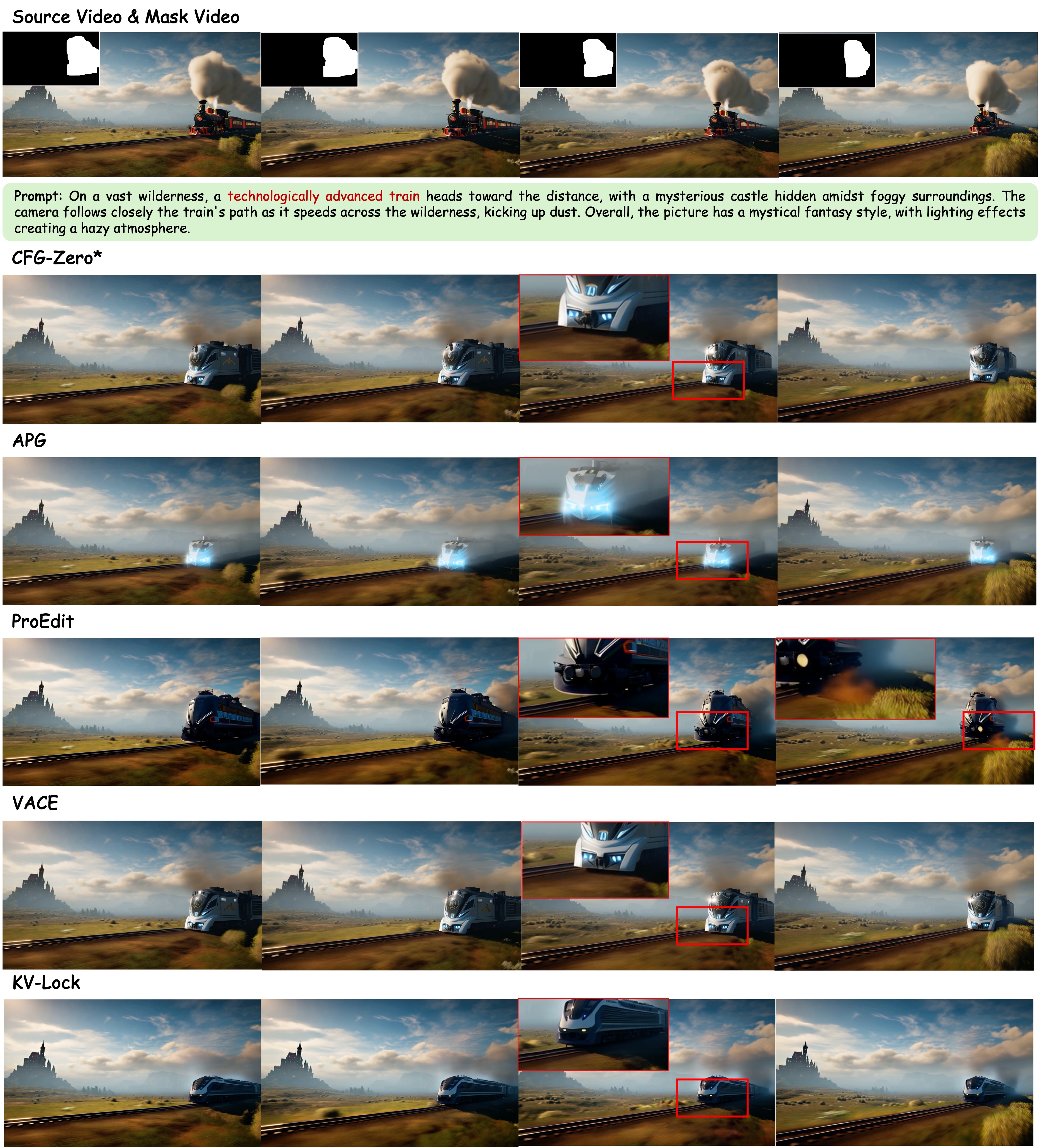}
    \caption{Video editing task without reference image. The task requires converting the steam locomotive in the original video into a technologically advanced train. As can be observed, the trains generated by CFG-Zero*, APG, and VACE suffer from obvious derailment. The train lights in the APG result distort the surrounding environment slightly. Moreover, the original long train is reduced to only two carriages in the outputs of CFG-Zero* and VACE. ProEdit produces unnatural connections between carriages and the distant background, as well as unrealistic ground dust. Only our method generates a properly scaled train without derailment, yielding natural and stable videos.}
    \label{fig:sup1}
\end{figure*}

\begin{figure*}
    \centering
    \includegraphics[width=\linewidth]{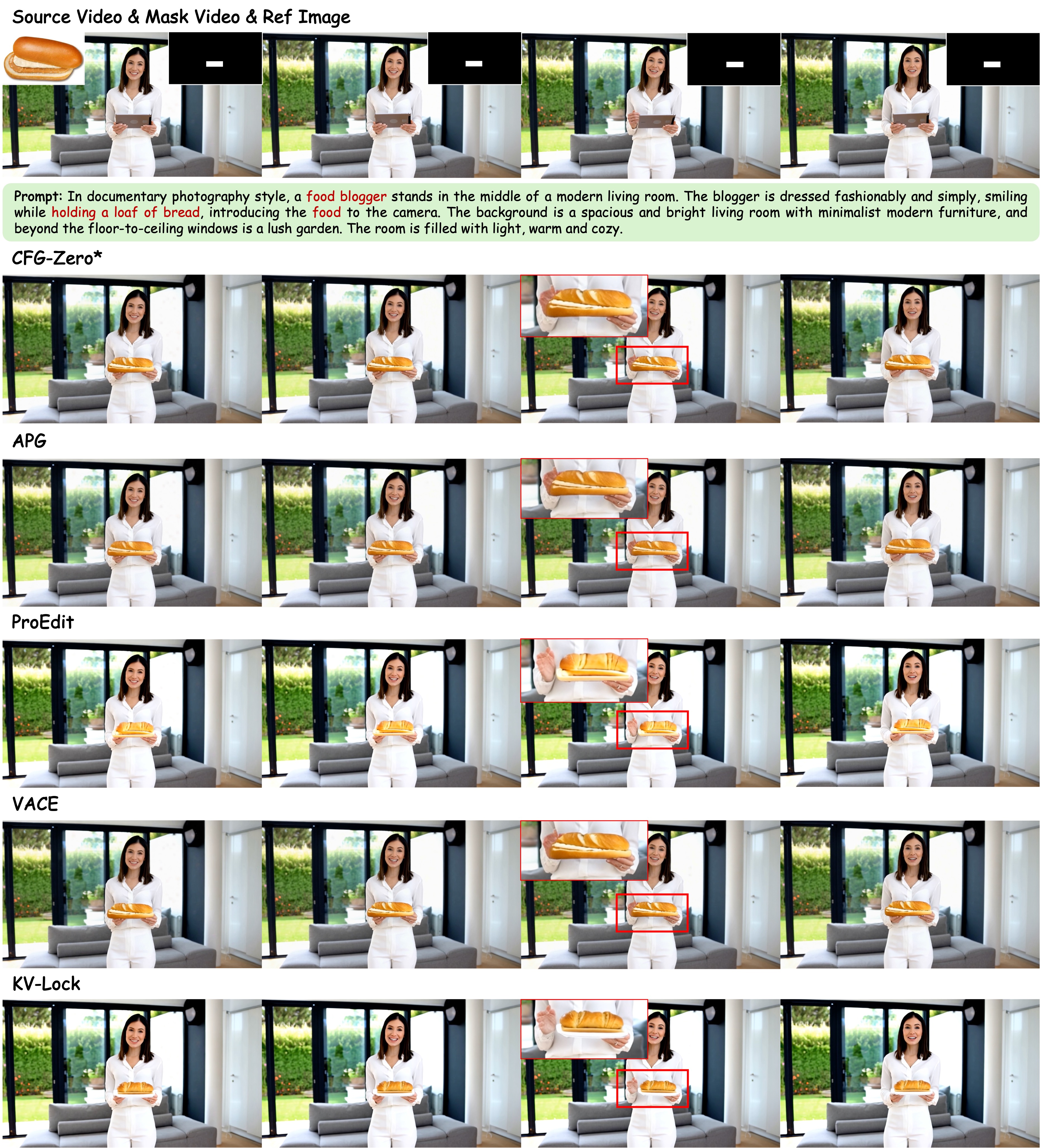}
    \caption{Video editing task with reference image. This task requires replacing the tablet held by the vlogger in the video with a piece of bread. Notably, the subject raises her right hand while presenting, leaving only her left hand to hold the object. As observed, CFG-Zero*, APG, and VACE directly replace the tablet with bread. However, physically, the bread cannot maintain such a posture when the right hand is raised; it would fall or at least tilt, making these results unrealistic. In contrast, ProEdit and our method introduce a tray under the bread to improve physical plausibility. Furthermore, the bread generated by ProEdit exhibits inconsistent brightness and saturation with the surrounding environment, while our method achieves both physical realism and environmental consistency.}
    \label{fig:sup2}
\end{figure*}

\begin{figure*}
    \centering
    \includegraphics[width=\linewidth]{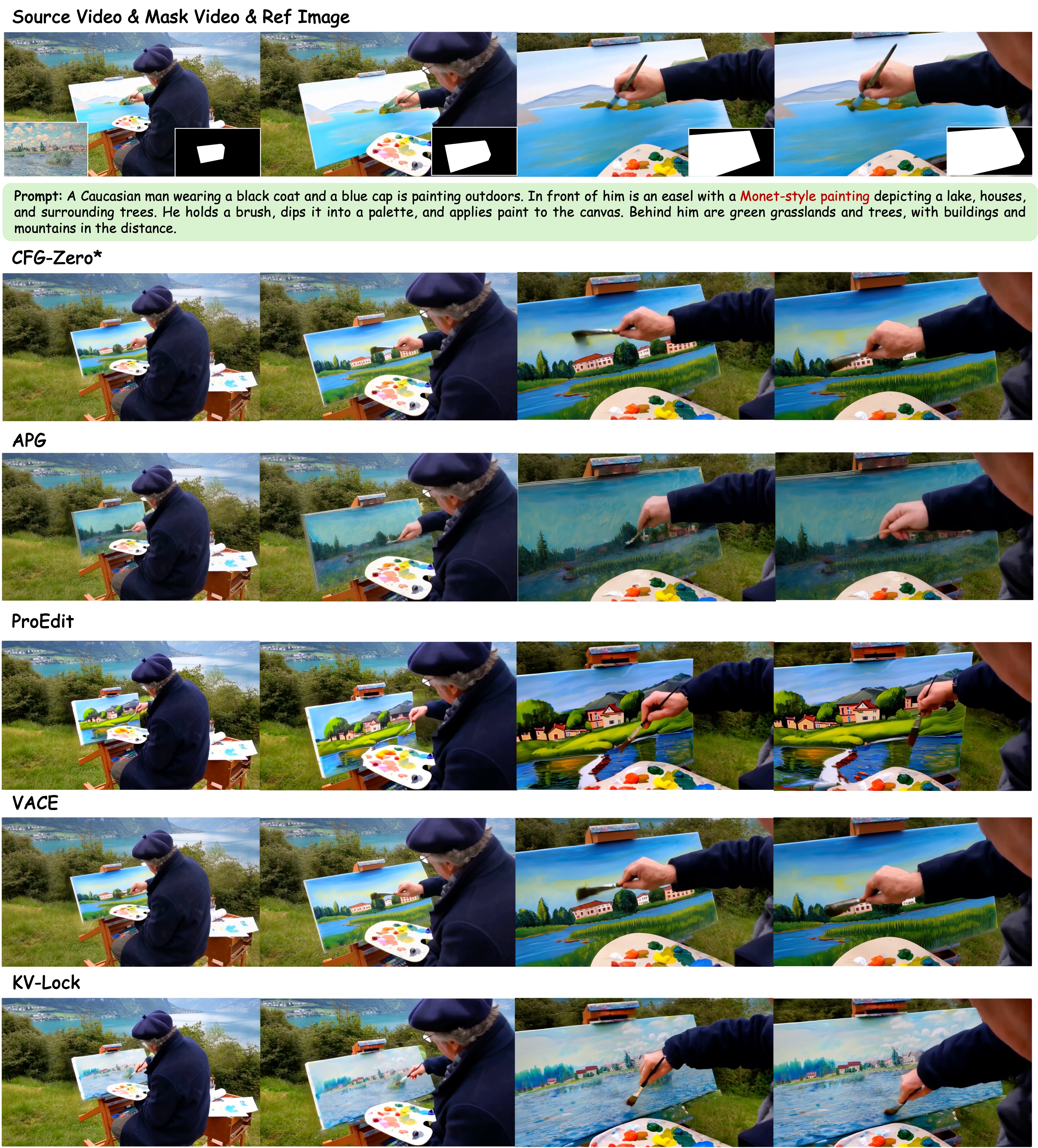}
    \caption{Video editing task with reference image. This task aims to replace the content on the canvas with Monet’s painting from the reference image. Due to interference from the surrounding context and the color information on the palette, none of the methods can directly and precisely transfer Monet’s painting onto the canvas. However, the results of CFG-Zero*, ProEdit, and VACE exhibit excessively high saturation and deviate completely from the style of Monet’s work, as the model directly colors the output based on the surrounding environment. The result of APG is overly dark in tone; although it attempts to approximate the target style, it still fails to capture the characteristics of the reference image. Only our method preserves the composition of Monet’s painting in the reference image. The waves, clouds, distant trees, and houses are clearly visible, and the overall brushwork is the closest to the target. The slight tonal difference is a reasonable adjustment made by the model to adapt to the ambient lighting.}
    \label{fig:sup3}
\end{figure*}

We present more experimental results in Figure \ref{fig:sup1} to Figure \ref{fig:sup3}, including video editing with and without reference images.
The results demonstrate that our method achieves superior performance in terms of visual quality, physical plausibility, and background consistency of the generated videos.

\end{document}